# Twitter-Based Gender Recognition Using Transformers

Zahra Movahedi Nia[1,2], Ali Ahmadi[3,4], Bruce Mellado[1,5], Jianhong Wu[1,2], James Orbinski[1,6], Ali Asgary[1,4], Jude Dzevela Kong[1,2,†]

**Abstract**—Social media contains useful information about people and the society that could help advance research in many different areas (e.g. by applying opinion mining, emotion/sentiment analysis, and statistical analysis) such as business and finance, health, socio-economic inequality and gender vulnerability. User demographics provide rich information that could help study the subject further. However, user demographics such as gender are considered private and are not freely available. In this study, we propose a model based on transformers to predict the user's gender from their images and tweets. We fine-tune a model based on Vision Transformers (ViT) to stratify female and male images. Next, we fine-tune another model based on Bidirectional Encoders Representations from Transformers (BERT) to recognize the user's gender by their tweets. This is highly beneficial, because not all users provide an image that indicates their gender. The gender of such users could be detected from their tweets. The combination model improves the accuracy of image and text classification models by 6.98% and 4.43%, respectively. This shows that the image and text classification models are capable of complementing each other by providing additional information to one another. We apply our method to the PAN-2018 dataset, and obtain an accuracy of 85.52%.

**Index Terms**— BERT, gender recognition, social media, transformers, ViT

———————————— ◆ ————————————

## 1 INTRODUCTION

PEOPLE are progressively becoming active in social media, sharing their thoughts, beliefs, concerns, and experiences. Consequently, a huge amount of useful information is produced that can help solve many problems in different areas such as economics [1], marketing [2], public safety and policy [3, 4], healthcare [5, 6], and gender vulnerability [30, 31]. User demographics provide social media-based research with essential information that can help study the issue from diverse perspectives. However, on most social media platforms, user information such as gender is considered private and therefore, is not freely available. As a result, user and author profiling has become a topic of great interest, recently.

By definition, author profiling is the process of using texts written by users for demographic extraction. Although many previous studies have focused on finding user information such as gender from text data [7-11], very few of them have considered using images. Combining image and text classification methods for finding users' genders can significantly increase the classification accuracy [22, 27]. In this paper, we propose a multimodal approach to find social-media users' gender by combining text and image processing and adapting transformers.

Transformers are novel deep learning models that use a self-attention mechanism to identify and learn significant parts of content [12]. The attention mechanism is a technique that is capable of enhancing and highlighting important parts of the content while downgrading other parts [13]. Self-attention is an attention mechanism that finds important tokens and their relations by comparing content with itself [14]. A token is usually a single word in Natural Language Processing (NLP), and a group of pixels processed together in computer vision. Since transformers can process tokens sequentially, they are suitable for both text and image processing [15, 16].

Transformers were initially used for NLP, and later on for computer vision. Before transformers, Recurrent Neural Network (RNN) models such as Long-Short Term Memory (LSTM) and Gated Recurrent Units (GRUs) with added attention layers on top of them were commonly used for NLP. In 2017, transformers were introduced by keeping the attention layer and dropping the RNN part to speed up the training process. Recently, transformers have been used and performed very well in image recognition. Bidirectional Encoder Representations from Transformers (BERT) [17] and Vision Transformers (ViT) [18] are models that are built using transformers and respectively trained for text and image classification.

BERT, which has become very popular for NLP lately, was first developed in 2018, solely from transformers. It is presented in two different modes, BERT$_{BASE}$ and BERT$_{LARGE}$ which respectively include twelve layers of transformers with twelve-headed bidirectional self-attention and twenty-four layers of transformers with sixteen-headed bidirectional self-attention. Both models have been trained in an unsupervised manner for language modelling and next-sentence prediction, using a large corpus gathered from books and Wikipedia pages.

[1]Africa-Canada Artificial Intelligence and Data Innovation Consortium (ACADIC)
[2]Laboratory for Industrial and Applied Mathematics, York University, Canada
[3]K.N. Toosi University, Faculty of Computer Engineering, Tehran, Iran.
[4]Advanced Disaster, Emergency and Rapid-response Simulation (ADER-SIM), York University, Toronto, Ontario, Canada
[5]School of Physics, Institute for Collider Particle Physics, University of Witwatersrand, Johannesburg, South Africa
[6]Dahdaleh Institute for Global Health Research, York University, Canada
[†]Corresponding Author: jdkong@york.ca

This time consuming computationally-expensive pre-training phase resulted in learning contextual embeddings for tokens i.e. words, by BERT. BERT can be fine-tuned to perform different NLP tasks such as question answering and language understanding, in a supervised manner. This can be extremely favourable to NLP developers since it performs very well with less labelled data.

ViT was developed in 2020, and published in 2021 by researchers from Google's Brain Team [29]. Since finding the relation between pixels is prohibitively complex in terms of memory and computation, ViT divides an image into 16x16 pixel sections for processing. Thus, a token is a 16x16 pixel piece of an image in ViT. Next, a learnable embedding vector is assigned to each token and along with positional embeddings are fed into a transformer architecture. Three different models are defined and trained for ViT, namely, ViT-Base, ViT-Large, and ViT-Huge, which respectively, include twelve layers of transformers with twelve-head self-attention; twenty-four layers of transformers with sixteen-head self-attention; and thirty-two layers of transformers with sixteen-head self-attention. The models have been pre-trained for image classification on different datasets including ImageNet, ImageNet-21k, and JFT-300M and have had up to 99.74% accuracy.

We labelled profile images for gender recognition and fine-tuned ViT based on that. Next, we applied our model to the PAN-2018 dataset [28] to predict users' gender using ten different images posted by a user. In addition, we fine-tuned a BERT model for text-based gender recognition using the PAN-2018 Twitter dataset. Eventually, we combined the image- and text-classification models using various machine learning algorithms, namely, Feedforward Neural Network (FNN), XGBoost, Random Forest (RF), Support Vector Machine (SVM), and Naive Bayes (NB).

The COVID-19 pandemic has exacerbated global socioeconomic inequalities, revealing how crises affect people differently according to their gender in troubling patterns which do not bode well for future resilience. Integrating governance at widening levels, and mitigating the limited economic options of women, are two examples of systematic challenges which require attention for human futurity – but in many cases, even the data required to document and understand these challenges is not available. This paper addresses these systematic imperatives by providing a framework that can help us to identify the elements of promising emergent governance frameworks to address local and global-scale socio-ecological challenges that disproportionately impact women.

In the following, Section 2 includes the literature review, Section 3 and 4 present our proposed method and numerical results, respectively, and Section 5 provides conclusion and future work, followed by a discussion in Section 6.

## 2 LITURATURE REVIEW

Author profiling has been practiced using different approaches [7-11]. Vashisth and Meehan [7] used different NLP methods for gender detection using Tweets, including Bag of Words (BoW) created with Term Frequency-Inverse Document Frequency (TF-IDF), word embeddings using W2Vec and GloVe embeddings, logistic regression, Support Vector Machine (SVM), and Naïve Bayes. They concluded that word embeddings have the highest performance for gender recognition. Ikae and Savoy [8] compared different machine learning methods for gender detection using tweets including logistic regression, decision tree, k-nearest neighbors (KNN), SVM, Naïve Bayes, neural networks, and random forest on seven different datasets. They concluded that neural networks and random forest perform best among the different approaches. Authors in [10] used n-grams as well as unigrams to tokenize sentences. They applied five different machine learning algorithms, Naive Bayes, Sequential Minimal Optimization (SMO), logistic regression, random forest, and j48 on text for gender recognition, and found that a combination of 1- to 4-grams with SMO produces the best accuracy.

The studies mentioned above, have only used text for gender recognition and have not considered image data. Authors in [19] were the first to use profile images for gender detection. They stacked different approaches, namely, Microsoft Discussion Graph Tool (DGT) using the username of the users, Face++ using their profile images, and SVMLight using their tweets. However, they combined pre-existing methods and did not train or fine-tune any model. In [20], VGG, a well-known image recognition model based on Convolutional Neural Networks (CNN) has been fine-tuned for gender detection of Twitter users. In [21], text and image have been used for predicting the gender of Twitter users. In the image classification method, a CNN is trained for gender recognition. The text classification method includes applying TF-IDF to the hashtags, and using Latent Dirichlet Allocation (LDA) to find the topics that the user is interested in. The results show that the combined method has higher accuracy.

Some studies have focused on image classification techniques for gender recognition. For example, authors in [23] propose a method for gender detection using images. First, they use CNN for feature extraction. Next, they apply a self-joint attention model for feature fusion. Finally, they use two fully connected neural network layers with ReLu and softmax activation functions, and one average pooling layer to predict the gender. In [24], a method using gated residual attention networks has been proposed for gender recognition using images and tested on five different datasets. In [25], different CNNs are trained for gender recognition using different methods such as KNN, Decision Tree, SVM, and softmax for feature extraction. The results of the CNN methods are combined by majority voting to increase the accuracy. Authors in [9] used posts, comments, and replies on facebook for gender recognition. They compared BERT with different machine learning and deep learning algorithms such as Naïve Bayes, Naïve Bayes Multinomial, SVM, decision tree, random forest, KNN, Recurrent Neural Networks (RNN), and Convolutional Neural Networks (CNN). The results show that BERT has the highest performance among the different methods.

Some studies have combined both text and image classification models and employed transformers for gender recognition. In [22], a multimodal approach using both text and image is proposed for the gender detection of Twitter users. The text classification part uses BERTBASE and the image classification part uses EfficientNet, a CNN-based approach for image recognition. The two methods are then combined to gain a higher accuracy. In [11], the gender of Twitter users has been predicted using their names, descriptions, tweets, and profile colours. SVM, BERT, and BLSTM have been applied to user descriptions, and BERT has performed better compared to SVM and BLSTM. Next, the different approaches are combined to improve the accuracy.

Methods mentioned above have used transformers for text classification for author profiling, and have found that BERT has higher performance compared to other methods. However, BERT has not been enhanced with image recognition using transformers for demographic information extraction. In this paper, we improve the performance and accuracy of text classification by combining BERT with image classification using ViT for gender recognition of Twitter users.

## 3 METHODOLOGY

Deep learning models such as transformers are advantageous to other machine learning models only when a large dataset is fed to them. Oftentimes, there is a need to employ supervised learning, particularly when labelled data is not available or it is very limited. In cases where a great amount of data is not accessible, fine-tuning a pre-trained deep learning model can help us find the desired accuracy. To this end, we have fine-tuned two powerful models based on transformers, namely, BERT and ViT which are respectively used for text and image classification to find Twitter users' gender. We applied our method on PAN-18 and compared text and image classification models with their combination in terms of accuracy, precision, recall, and Fi-score, and found that their combination is superior in all of the metrics, and results in high accuracy.

### 3.1 Fine-Tuning ViT

We retrieved geotagged tweets from Canada and South Africa using Twitter's Application Programming Interface (API) academic researcher account. The users were manually labelled into three classes, namely, female, male, and unknown (the class unknown here does not imply gender-neutral), based on their Twitter profile images. The unknown class included images that did not suggest a gender, because they did not contain human faces; or involved only children, or couples, or a group of women and men. We recognize that binary classification of gender into female and male excludes other forms of gender representations. Our train and test datasets included 467 and 161 female, 468 and 156 male, and 451 and 150 unknown images that are pretty balanced (See figure 1). These datasets were used to fine-tune ViT for gender classification.

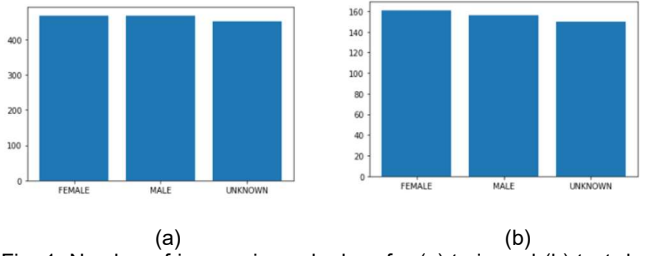

Fig. 1. Number of images in each class for (a) train and (b) test datasets

Next, we applied our model to the PAN-18 dataset [28]. This dataset includes text and images of English, Spanish, and Arabic-speaking users. There are ten different images available for each user. Hoping that most users do post pictures of themselves, we used our trained model to classify the users based on gender by taking the ten different images that they had posted. We combined the result of the ten images of each user using different machine learning algorithms, i.e. FNN (three neurons in the hidden and two in the output layers), XGBoost, RF, SVM, and NB, and achieved the highest accuracy with RF (see figure 2).

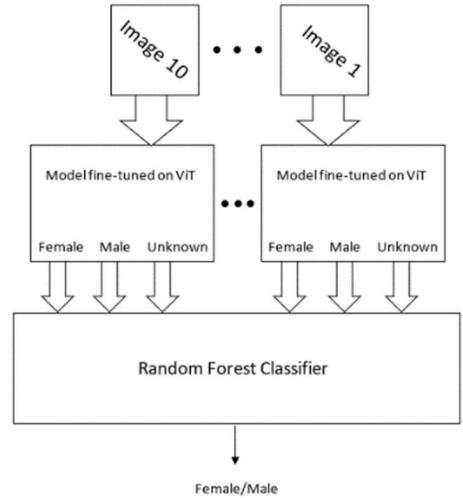

Fig. 2. The image classification model for gender recognition

### 3.2 Fine-Tuning BERT

Some Twitter users may not have a suitable image for detecting their gender. However, we are able to retrieve the tweets of most Twitter users. Therefore, training a text classification model for gender recognition could help extract the gender of more users and increase the performance of the model. PAN-18 dataset includes 3000 and 1900 users for training and testing. Half of the users are female and half male, and the dataset is completely balanced. In the PAN-18 dataset, one hundred tweets are available for each user, which we used to fine-tune the BERT model for gender recognition. We tried to identify the gender of users by a single tweet, however the accuracy of the model did not exceed 59%. Thus, since the num-ber of tokens for BERT cannot be more than 512, we concatenated ten randomly chosen tweets and found ten concatenated tweets for each user. We trained a BERT model using the concatenated tweets, and combined the ten different concatenated tweets of each user using various machine learning algorithms, i.e. FNN, XGBoost, RF, SVM, and NB, and found 81.89% accuracy with FNN. As shown in figure 3, the FNN model consisted of a hidden and an output layer of three and two neurons. Adding more layers and neurons did not increase the accuracy of the model.

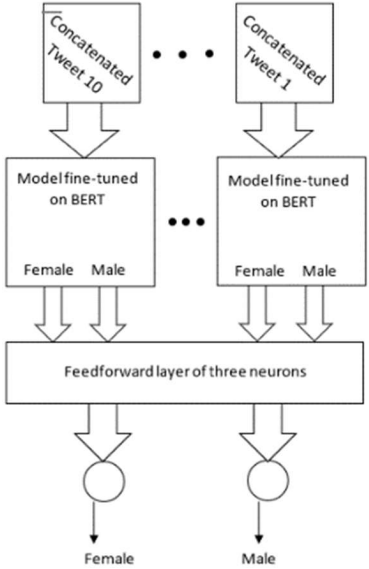

Fig. 3. The text classification model for gender recognition

### 3.3 Combining Image and Text Classification

For each user of the PAN-18 dataset, ten different images, and ten concatenated tweets were extracted. Each image went through an image classification model fine-tuned on ViT, and each concatenated tweet went through a text classification model fine-tuned on BERT. The image classification model has three outputs, i.e. female, male, and unknown, and the text classification model has two outputs, i.e. female and male. Therefore, fifty outputs for each user were obtained, thirty of which came from the image-classification model, and twenty from the text-classification model (see figure 4). We placed a feedforward neural network on top of these fifty outputs to combine the models' output. The feedforward network consisted of two layers. The hidden and output layers included ten and two neurons, respectively. Our code is available at [32].

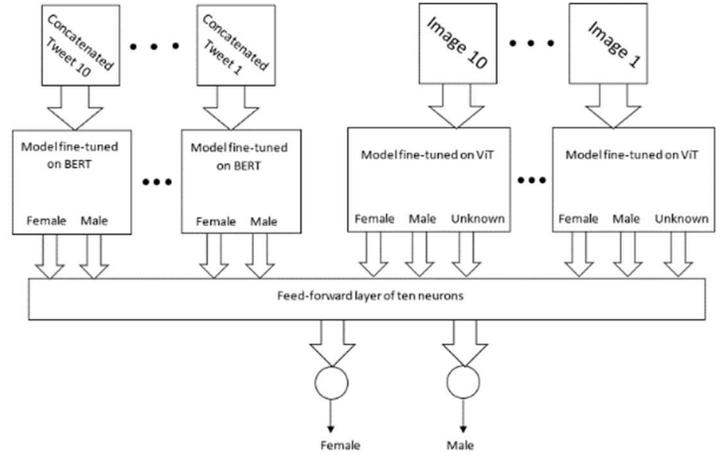

Fig. 4. The multimodal approach for gender recognition

## 4 RESULTS

### 4.1 Gender Recognition with Fine-Tuning ViT

Figure 5 shows the confusion matrix of the model fine-tuned on ViT using the dataset gathered from Twitter. The result implies that our model was able to classify the tweets with more than 88% accuracy.

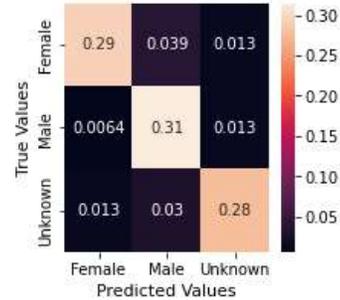

Fig. 5. The confusion matrix of the model fine-tuned on ViT using our Twitter dataset

Next, we used our model to classify the images of PAN-18. The results of the ten images were combined using different machine learning models. Table 1 compares the model trained on the Twitter dataset and the model trained on the ten images of a user from the PAN-18 dataset. According to Table 1, the models trained on our Twitter dataset and the ten images of a user from the PAN-18 dataset have high precision, recall, and F1-score for all of their classes, i.e. female, male, and unknown. Precision indicates the percentage of the correctly classified items detected for a particular class, and recall indicates the percentage of the items from a particular class that were actually detected. High precision and recall for all the classes indicate that the models can distinguish between all the classes pretty well.

TABLE 1
METRICS OF THE MODEL TRAINED WITH OUR TWITTER DATASET

|  |  | Model trained with our Twitter dataset |
|---|---|---|
| Accuracy |  | 88.65% |
| Precision | Female | 0.94% |
|  | Male | 0.82% |
|  | Unknown | 0.92% |
| Recall | Female | 0.85% |
|  | Male | 0.94% |
|  | Unknown | 0.87% |
| F1-Score | Female | 0.89% |
|  | Male | 0.88% |
|  | Unknown | 0.89% |

Figure 6 shows the confusion matrix of the gender identification model that combines ten different images of a user using RF. As shown in figure 6 the final accuracy of the image classification model is approximately 80%.

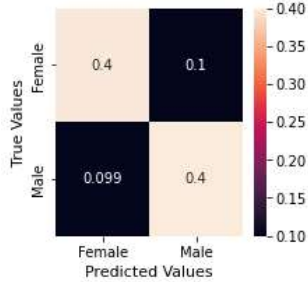

Fig. 6. The confusion matrix of the final model combining ten images of a user by RF

Table 2 compares the accuracy of different machine learning algorithms used for combining the images. These results demonstrate that the highest accuracy for combining the ten different images of a user from the PAN-18 dataset was obtained using a RF classifier.

TABLE 2
METRICS OF THE MODELS TRAINED FOR COMBINING IMAGES (IN PERCENTAGE)

|  |  | FNN | XGB | RF | SVM | NB |
|---|---|---|---|---|---|---|
| Accuracy |  | 77.68 | 77.89 | **79.94** | 77.84 | 79.11 |
| Precision | Female | 79 | 80 | 80 | 79 | 81 |
|  | Male | 76 | 76 | 80 | 77 | 78 |
| Recall | Female | 75 | 75 | 79 | 76 | 76 |
|  | Male | 81 | 81 | 80 | 80 | 82 |
| F1-Score | Female | 77 | 77 | 80 | 77 | 78 |
|  | Male | 78 | 79 | 80 | 78 | 80 |

### 4.2 Gender Recognition with Fine-Tuning BERT

Figure 7 shows the confusion matrix of the model fine-tuned on BERT for gender recognition using tweets. These results show that the genders are identified correctly with almost 82% accuracy. Moreover, the confusion between females and males is very low which indicates that the two classes are separated pretty well, and the precision and recall are very high.

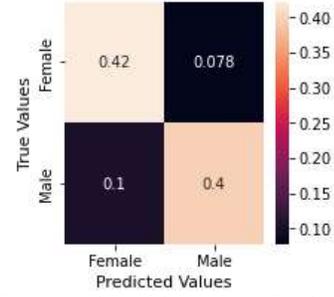

Fig. 7. The confusion matrix of the model fine-tuned on BERT for gender recognition using the PAN-18 dataset

Table 3 compares different machine learning models used for combining the ten concatenated tweets from the PAN-18 dataset, in terms of accuracy. According to these findings, combining the ten concatenated tweets using the FNN, results in the highest accuracy amongst the machine learning algorithms used.

TABLE 3
METRICS OF THE MODELS TRAINED FOR COMBINING TEXT (IN PERCENTAGE)

|  |  | FNN | XGB | RF | SVM | NB |
|---|---|---|---|---|---|---|
| Accuracy |  | **81.89** | 80.89 | 81.47 | 81.78 | 81 |
| Precision | Female | 80 | 80 | 81 | 80 | 80 |
|  | Male | 84 | 82 | 82 | 83 | 82 |
| Recall | Female | 84 | 82 | 83 | 84 | 83 |
|  | Male | 79 | 80 | 80 | 79 | 79 |
| F1-Score | Female | 82 | 81 | 82 | 82 | 81 |
|  | Male | 81 | 81 | 81 | 81 | 81 |

### 4.3 Gender Recognition with Fine-Tuning BERT

Figure 8 shows the confusion matrix of the final model which combines text- and image-based models for gender recognition using the PAN-18 dataset. The results show that the final model has more than 85% accuracy.

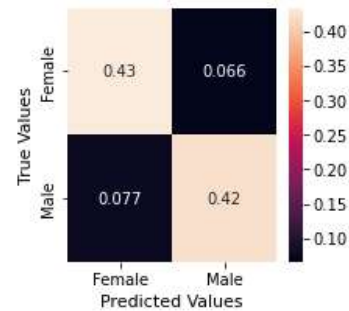

Fig. 8. The confusion matrix of the final model combining image- and text-based models for gender recognition using the PAN-18 dataset

Table 4 compares the different machine learning algorithms used for combining text- and image-based gender recognition models using the PAN-18 dataset. Table 5 compares the metrics of the image-, text-, and image+text-based models for gender recognition using the PAN-18 dataset.

TABLE 4
METRICS OF THE MODELS TRAINED FOR COMBINING TEXT (IN PERCENTAGE)

|  |  | FNN | XGB | RF | SVM | NB |
|---|---|---|---|---|---|---|
| Accuracy |  | **85.52** | 85.47 | 82.73 | 85.48 | 82 |
| Precision | Female | 85 | 85 | 81 | 85 | 81 |
|  | Male | 86 | 85 | 82 | 87 | 84 |
| Recall | Female | 87 | 85 | 83 | 87 | 85 |
|  | Male | 85 | 85 | 81 | 84 | 80 |
| F1-Score | Female | 86 | 85 | 82 | 86 | 83 |
|  | Male | 85 | 85 | 82 | 85 | 82 |

Table 5 compares different metrics of the image-, text-, and image+text-based models for gender recognition using the PAN-18 dataset.

TABLE 5
METRICS OF IMAGE-, TEXT-, AND IMAGE+TEXT-BASED MODELS FOR GENDER RECOGNITION USING THE PAN-18 DATASET (IN PERCENTAGE)

|  |  | Image | Text | Image+Text |
|---|---|---|---|---|
| Accuracy |  | 79.94 | 81.89 | **85.52** |
| Precision | Female | 80 | 80 | **85** |
|  | Male | 80 | 84 | **86** |
| Recall | Female | 79 | 84 | **87** |
|  | Male | 80 | 79 | **85** |
| F1-Score | Female | 80 | 82 | **86** |
|  | Male | 80 | 81 | **85** |

According to Table 4, FNN provides the highest accuracy for combining text- and image-based gender classification models using the PAN-18 dataset. Adding more layers and neurons to the FNN did not increase the accuracy of the model. According to Table 5, all the metrics of the combined model are higher than that of the image- and text-based gender recognition models, which shows the superiority of the combined model. By combining image- and text-based gender recognition models, we obtained an accuracy that is 6.98% higher than that of the image-based gender recognition model and 4.43% higher than that of the text-based gender recognition model. This indicates that text and image carry additional information relative to one another and can complement each other, and increase the accuracy of the gender classification model.

## 5 DISCUSSION

Demographics of social media users are beneficial for research and applications in health, socio-economic inequalities and gender vulnerability. However, such information is not usually and freely available. In this paper, we have designed a framework for detecting the gender of Twitter users. To this end, we first, fine-tuned a model based on ViT for gender classification. With a small amount of data, we obtain an accuracy of 88.65%. We applied our framework to the PAN-18 dataset and fine-tuned a text-classification model on BERT using the PAN-18 dataset, as well. We then combined the two models using a feedforward neural network and found that the image- and text-based models were able to complement each other. The accuracy of the combined model was equal to 85.52% which is significantly higher than that of each of the models.

Future studies could build on our work by using other user information, such as descriptions, media posts, comments, and likes. Moreover, recognizing other user demographics such as age, and ethnicity using transformers could be further investigated. In addition, heuristic methods for identifying user demographics when images are blurry, have low quality, are partially viewed, or when people are wearing masks, or sunglasses can be studied for higher accuracy and better performance.

## 6 CONCLUSION

During periods of upheaval, women are usually at greater risk from the adverse effects-, and potential losses incurred by these external stressors. They are also the slowest to recover from such emergencies. Integrating governance at widening levels, and mitigating the limited economic options of women, are two examples of systemic challenges which require attention for human futurity – but in many cases, even the data required to document and understand these challenges is not available. This paper addresses these systemic imperatives by providing a framework that can help us to identify the elements of promising emergent governance frameworks to address local and global-scale socio-economic challenges that disproportionately impact women.

### ACKNOWLEDGEMENT

This research is funded by Canada's International Development Research Centre (IDRC) and the Swedish International Development Cooperation Agency (SIDA) (Grant No. 109559-001).